\documentclass[lettersize,journal]{IEEEtran}
\IEEEoverridecommandlockouts
\usepackage{cite}
\usepackage{amsmath,amssymb,amsfonts}
\usepackage{svg}
\usepackage{dirtytalk}
\usepackage{algorithmic}
\usepackage{graphicx}
\usepackage{textcomp}
\usepackage{xcolor}
\usepackage{wrapfig}
\usepackage{multirow}
\usepackage{adjustbox}
\usepackage{mathcomp}
\usepackage{pdfpages}

\usepackage[inline]{enumitem}
\usepackage[breaklinks]{hyperref}
\usepackage[section]{placeins}

\begin{document}
\title{BlabberSeg: Real-Time Embedded \\ Open-Vocabulary Aerial Segmentation}

\author{
        Haechan Mark Bong$^{*}$,~\IEEEmembership{Student Member,~IEEE,}
        Ricardo de Azambuja$^{*,\dagger}$,~\IEEEmembership{Member,~IEEE,} \\
        Giovanni Beltrame$^{*}$,~\IEEEmembership{Senior Member,~IEEE}
	\thanks{$^{*}$MISTLab, \'Ecole Polytechnique Montr\'eal, Montr\'eal, Canada}
        \thanks{$^\dagger$Currently at Questat.ca (contributed to this work while at MISTLab).}
	\thanks{{\tt Contact: haechan.bong@polymtl.ca}}
        \thanks{This work was supported by the National Research Council Canada (NRC).}
}

\maketitle
\bstctlcite{IEEEexample:BSTcontrol}

\begin{abstract}
  Real-time aerial image segmentation plays an important role in the
  environmental perception of Uncrewed Aerial Vehicles (UAVs). We introduce
  BlabberSeg, an optimized Vision-Language Model built on CLIPSeg for on-board,
  real-time processing of aerial images by UAVs. BlabberSeg improves the
  efficiency of CLIPSeg by reusing prompt and model features, reducing
  computational overhead while achieving real-time open-vocabulary aerial
  segmentation. We validated BlabberSeg in a safe landing scenario using the
  Dynamic Open-Vocabulary Enhanced SafE-Landing with Intelligence (DOVESEI)
  framework, which uses visual servoing and open-vocabulary segmentation.
  BlabberSeg reduces computational costs significantly, with a speed increase of
  927.41\% (16.78 Hz) on a NVIDIA Jetson Orin AGX (64GB) compared with the
  original CLIPSeg (1.81Hz), achieving real-time aerial segmentation with
  negligible loss in accuracy (2.1\% as the ratio of the correctly segmented
  area with respect to CLIPSeg). BlabberSeg's source code is open and available
  online\footnote{\hyperlink{https://github.com/MISTLab/BlabberSeg}{https://github.com/MISTLab/BlabberSeg}}.
\end{abstract}


\section{Introduction}
\IEEEPARstart{U}{nmanned} Aerial Vehicles (UAV) are gaining prominent attention
from industry and academia, especially with studies focusing on Intelligent
UAVs~\cite{pal2023comprehensivereviewaienabledunmanned}. By mounting AI-capable
devices onto UAVs, it is possible to enable Cognitive Internet of Things
(CIoT)~\cite{fayaz2021cognitive} on UAVs and perform image processing and
segmentation tasks such as object detection and segmentation, which are
essential for the safe automation of flight operations.

With the rise of foundation models such as Large Language Models (LLMs) and
Vision Language Models (VLMs), there is an exponential growth of interest in
using these models to improve the automation of vehicles and
machines~\cite{gdm2024autort,huang2023embodied}. In the context of UAVs, Liu et
al.~\cite{liu2023remoteclip} and Bong et al.~\cite{bong2023peace} showcased
generative AI capabilities on aerial imaging using foundational models. Other
research efforts~\cite{rana2023sayplan,kira2022llmroboticspaperslist} focused on
performing specific tasks such as navigation, planning, exploration and object
detection using CIoT devices equipped with LLMs. Although promising results were
demonstrated, they share a common constraint, which is the requirement of a high
performance on-board computer. In particular, recent LLMs and
VLMs~\cite{openai2023gpt4,radford2021learning,touvron2023llama} require massive
computing power during training and segmentation. Due to the limited
computational capabilities of CIoT devices, most foundation models on CIoT
devices are not suitable for real-time decision making.

Numerous works attempted to optimize the efficiency of neural network
segmentation on CIoT devices, using various techniques such as model
compression~\cite{Niu_2020,wu2016quantized}, input
filtering~\cite{MLSYS2022_d59a1dc4,yuan2023infi}, and
reusing~\cite{DBLP:conf/percom/DroliaGTGN17,Xu_2018}. However, they are static
acceleration methods, which may lead to significant performance degradation when
used with foundation models. Dynamic approaches like early exit
strategies~\cite{10.1145/3534619,Leontiadis_2021} can reduce unnecessary
computations, but the substantial dimensions and deep layers of foundation
models result in heavyweight early-exit heads. Moreover, optimizing computation
efficiency through input data processing, including filtering and reusing do not
adequately address the persistent challenge of limited memory on edge-based CIoT
devices.

To mitigate these computational constraints, Yang et al.~\cite{yang2023edgefm}
proposed moving the computation of foundation models to edge infrastructure.
However, such approach is not practical on UAVs operating with unstable wireless
connections (forests, caves, etc.). Other approaches focus on Tiny Machine
Learning (TinyML), a relatively new concept that focuses on building lightweight
learning models to accommodate on-device operations. For example, the Machine
Learning Compilation (MLC) project~\cite{mlc-llm} uses memory planning and
quantization techniques to reduce computation latency, while Wang et
al.~\cite{10.1145/3552326.3587438} and Chen et al.~\cite{chen2023frugalgpt}
focus on reducing the model size. Although their results are promising, they are
built for text-dialogue generation, which is not suitable for UAV operations,
which are dependent to multi-modal sensors and camera. Knowledge-distillation
based models such as DIME-FM~\cite{sun2023dimefm},
FD-CLIP~\cite{wei2022contrastive}, VLKD~\cite{dai2022enabling}, and
Mobile-SAM~\cite{zhang2023faster} are capable of compressing multi-modal
foundation models to certain extent, but they are still heavyweight and yet to
be tested on real-time CIoT devices mounted on UAVs.


We propose BlabberSeg, an optimized and extended version of one of the most
common VLM segmentation models CLIPSeg~\cite{Luddecke_2022_CVPR}, which is based
on CLIP (Contrastive Language-Image Pretraining). In UAV operations, there are
often multiple detection and segmentation tasks to be performed. Using CLIPSeg,
prompting with a single target label performs much better than a prompt with all
the target labels due to accumulation of noise and disparity in cosine
similarity~\cite{bong2023dynamic}. A prompt per target label is not only
computationally expensive, but also not scalable, especially if the target class
is not fixed. Aerial images, especially at high altitudes (100m) typically have
low resolution~\cite{10411503}.


Within these constraints, BlabberSeg increases segmentation speed by reusing
prompt and image embeddings without losing the overall segmentation accuracy
(the ratio of the correctly segmented area over the ground truth (segmentation
of the original CLIPSeg)). Reusing of embeddings becomes exponentially more
significant with the number of prompts.

To evaluate BlabberSeg, we consider emergency UAV landing as a target
application. Many previous systems dealt with automatic UAV safe landing, but
they would mostly limit the maximum altitude to under
30m~\cite{Mittal2018VisionbasedAL,chatzikalymnios2022,7138988,rabah2018,6884813}
since higher altitude makes aerial segmentation very difficult, especially when
UAVs are moving. We evaluate BlabberSeg using Google Maps~\cite{GoogleMaps2023}
using DOVESEI~\cite{bong2023dynamic}, an emergency safe landing system that uses
CLIPSeg~\cite{Luddecke_2022_CVPR} for aerial safe landing zone (SLZ) detection
at an altitude of 100m. We used an NVIDIA Jetson Orin AGX~\cite{nvidia} as a
testing device throughout our experimentation. BlabberSeg drastically lowers
computational costs, delivering a 927.41\% increase in speed (16.78 Hz) on an
NVIDIA Jetson Orin AGX (64GB)~\cite{nvidia} compared to the original CLIPSeg
(1.81 Hz). This allows for real-time aerial segmentation while maintaining
minimal reductions in accuracy (2.1\%) and mean Intersection over Union (mIoU)
(9\%).


\section{Materials and Methods}
\begin{figure*}
\centerline{\includegraphics[width =1\linewidth, trim = 0 5.9cm 0 5.99cm, clip]{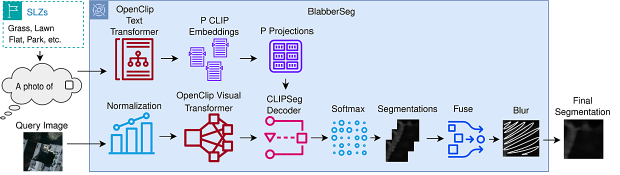}}
\caption{BlabberSeg Architecture: In this architecture, we harness OpenCLIP Visual Transformer's capability to cast CLIP into floating-point (FP) 16 and use FP16 converted CLIPSeg Decoder with reused prompt and positional embeddings to accelerate segmentation. P represents the number of prompts, which are the number of words that are selected to describe safe landing zones (e.g. Grass, Lawn, Flat, Park, etc.).}
\label{methods:blabberseg}
\end{figure*}

\subsection{Reusing for Efficiency}
BlabberSeg enables significant computational efficiency on CLIPSeg by reusing
features to achieve data processing suited for CIoT devices. In a UAV context,
input images and prompts are similar in content since the environment where the
UAV flies at a given time will stay the same. However, VLMs are designed to take
different images and prompts every time they are processed and BlabberSeg can
mitigate re-computation through reusing, which is well-suited for lightweight
UAV operations. In addition, our model allows multiple prompts on a single image
to target multi-label segmentation.

\subsubsection{Reusing Positional Embeddings}
The conventional approach with image segmentation involves recalculating
positional embeddings for every image, a process that can be computationally
resource-intensive. In our optimized approach, we pre-compute the rescaling of
positional embeddings, fixing it according to the input image size. While the
optimal image size for CLIPSeg is 352, our method accommodates smaller sizes,
thereby reducing the overall model size and improving the processing speed. More
specifically, our model treats the positional encoding as an \say{image} that
maintain the same number of patches (considering a fixed image size relative to
patch size). This interpolated positional encoding is calculated once and reused
for subsequent incoming images, eliminating the need for repetitive calculations
during segmentation. In cases where a new image has a different size, the
positional encoding is dynamically adjusted through shrinking or interpolation
to align with the pre-computed positional encoding.

\subsubsection{Optimization of CLIP Activations}
In the original CLIPSeg model, the CLIP image encoder and its activations are
invoked for each prompt, leading to redundant computations for real-time image
processing in UAVs. Our optimization addresses this inefficiency by running CLIP
activations only once per image, which notably reduces the computational
redundancy. Furthermore, we simplify the computation of CLIP activations to
calculate only the necessary components, avoiding inessential computations that
yield the same output as CLIP's entire image encoder. More specifically, we
reduced the original 12 layers used in CLIP to 10 layers by always using batch
size one and images with a fixed size, the tensor filled with zeros will be
always the same and thus they are reused. This streamlined approach enhances the
overall efficiency of the CLIPSeg model, ensuring that only essential
computations are performed during semantic segmentation.

\subsubsection{Pre-computing FiLM}
In the original CLIPSeg, the model could only reuse conditionals, but our
optimized version pre-computes Feature-wise Linear Modulation
(FiLM)~\cite{perez2017film} on conditionals within the decoder per prompt,
allowing for their efficient reuse. This advancement significantly reduces
computational overhead during the segmentation process, enhancing the model's
overall efficiency of CLIPSeg decoder, allowing for segmentation per prompt with
reduced computational complexity.

\subsubsection{Image Processing}
In order to facilitate pre and post image processing, input images are
normalized and output segmentations use softmax. In addition, each segmentation
generated per prompt are fused to generate a final segmentation with all prompts
for a single image. Then we apply blur to make segmentation smoother, just like the
DOVESEI~\cite{bong2023dynamic} system.

\subsubsection{Hardware Acceleration}
Other than reusing, we accelerated our model using TensorRT and ONNX runtime
input/output binding, taking advantage of efficient hardware acceleration. In
addition, our model integrated OpenCLIP instead of the original CLIP to address
precision-related challenges by casting our model with floating-point 16 (FP16).
The internal cast to FP32 in CLIP can disrupt compatibility with FP16, making it
challenging to operate with different precision. OpenCLIP provides a solution
that enables seamless casting of the entire model to FP16, enhancing flexibility
without compromising precision. Our model is then converted to two ONNX models
(activations and CLIPSeg decoder) which are then converted to TensorRT engines
to further accelerate segmentation time.

In summary, BlabberSeg enables the advantage of using multiple prompts per image
in the segmentation process, focusing on reusing essential components to save
computations in on-device aerial imaging. These optimizations collectively
contribute to a more efficient CLIPSeg model, making it adaptable to a broader
range of image sizes while maintaining high performance in semantic segmentation
tasks. The proposed model offer a balance between computational efficiency and
model accuracy, addressing the computational challenges associated with
on-device image segmentation in UAVs.

\subsection{Experiment Setup}
Our experiments consist of demonstrating how each reused features increase
computational efficiency. Our tests are done in a complex UAV context, which is
the use-case of urgent safe landing.

\subsubsection{CIoT Device Setup}
In order to facilitate reproducible experiments, all tests are done in NVIDIA
Jetson devices~\cite{nvidia} and Raspberry Pi 5~\cite{rpi5}. Our main CIoT
device for testing was NVIDIA Jetson AGX Orin 64GB~\cite{nvidia}. To demonstrate
the efficiency of BlabberSeg in less powerful CIoT devices, we also measured
performance efficiency using AGX Xavier~\cite{nvidia} 64GB, Raspberry Pi
5~\cite{rpi5} (CPU only), and emulated NVIDIA Jetson Orin NX 16GB~\cite{nvidia}
and NVIDIA Jetson Orin Nano 8GB~\cite{nvidia}.

\subsubsection{DOVESEI Architecture}
\begin{figure}[htbp]
\centerline{\includegraphics[width=1.0\linewidth]{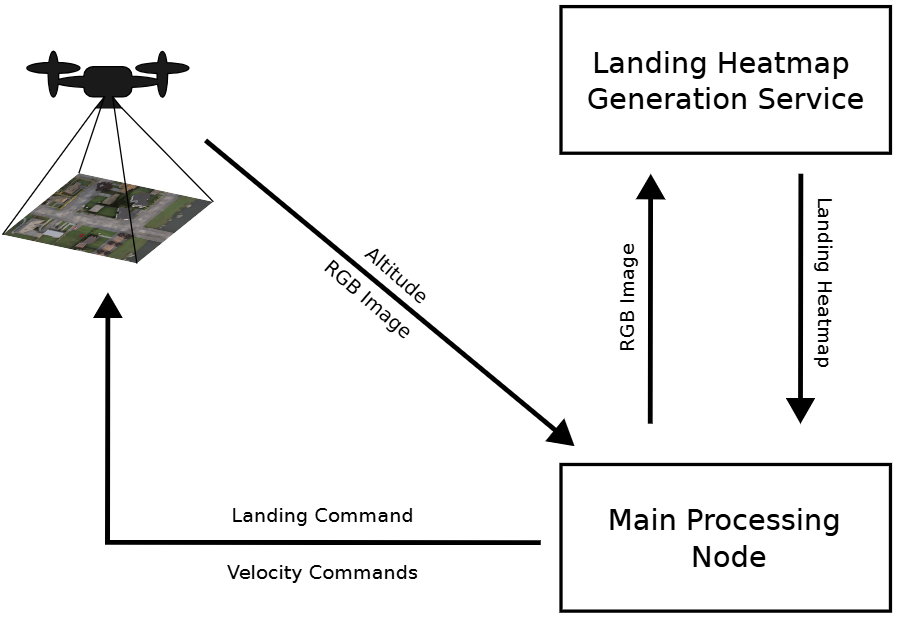}}
\caption{DOVESEI~\cite{bong2023dynamic} was implemented in ROS~2 and it is composed of three main blocks: UAV (flight controller, sensors), landing heatmap generation (receives an RGB image and produces a heatmap of the best places to land), and main processing node (orchestrates the data exchange with the UAV, sends velocity commands).}
\label{methods:main}
\end{figure}
To validate the robustness of our model in a complex and realistic environment,
DOVESEI~\cite{bong2023dynamic} system was used to simulate an urgent UAV safe
landing scenario with satellite images of Paris (Google Maps~\cite{GoogleMaps2023}). Using this system, we were
able to test our work in high-altitude computing (HAC) and low-altitude
computing (LAC) by starting at 100m and descending to 20m. Its system
architecture is setup using ROS~2~\cite{Thomas2014} package and composed of two
principal interconnected components: Landing Heatmap Generation Service and Main
Processing Node (Fig.~\ref{methods:main}).

A RGB image and textual prompts serve as inputs to the heatmap generator, which
employs BlabberSeg to generate a comprehensive heatmap in a zero-shot manner.
This heatmap provides essential insights into optimal landing positions,
consistently referred to as \say{best} within the given image frame context.

The main node assumes high-level control over the entire system and establishes
direct connections with the UAV flight controller. Its functionality encompasses
three core components: the main state machine, post-processing of raw heatmaps,
and dynamic focus.

The state machine governs the dynamic behavior of our system, with its primary
states being:
\begin{enumerate*}[label=\textbf{\roman*})] 
	\item Searching: Coarse search for a landing spot from a safe (collision free) altitude.
	\item Aiming: Refined search to better align the UAV with a safe landing spot.
    \item Landing: Descend while checking for dynamic obstacles.
    \item Waiting: Stop and wait if any obstacles were detected after it started landing.
    \item Climbing: Climb back to the safe altitude if the waiting phase triggered a failure.
    \item Restarting: Restart the coarse search by moving to a new starting position.
\end{enumerate*}

The dynamic focus oversees the degree to which the Raw Heatmap Post-processing
module processes the raw heatmap. It selectively \say{focuses} by applying a
binary mask that encompasses specific regions of the input, prioritizing the
most crucial areas based on the current system state and its operation.

\section{Results and Discussions}
In this section, we will refer to TensorRT engine and ONNX runtime input/output
binding as TensorRT and Input/Output Binding, respectively.

Our primary objective was to increase CLIPSeg segmentation speed to achieve
real-time SLZ detection, ideally optimize throughput to 10Hz. Decrease in
segmentation accuracy was expected as an exchange for the gain of segmentation
speed. However, we had to ensure that the drop in accuracy is minor in order to
be deployed in emergency safe landing applications. Using 500 aerial
images~\cite{GoogleMaps2023}, we compared the mean segmentation accuracy and
Mean Intersection over Union (mIoU) between the original CLIPSeg (considered as
a ground truth) and optimized CLIPSeg models. Our threshold for
acceptable mean accuracy and mIoU was set to 90 to ensure that our optimization
methods will not hinder segmentation performance. Note that our objective in
this experiment was not to measure the segmentation quality, we aim to minimize
the change on segmentation results when optimization was applied. The various
optimizations techniques applied are listed in Table~\ref{comparision}. We also
tried to optimize HF (Hugging Face) version of CLIPSeg, but we realized that
their segmentation speeds are much slower than the original CLIPSeg
(Table~\ref{comparision}).
 
\begin{figure}[htbp]
\centerline{\includegraphics[width =1\linewidth]{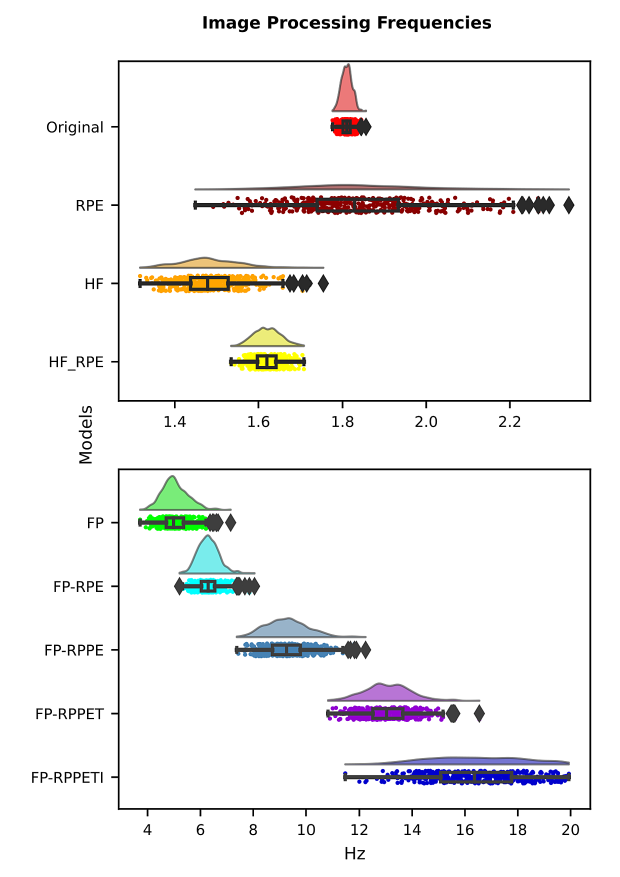}}
\caption{Increase in Frequency Through Optimization. \\\\
Model Legend: \\
Original: CLIPSeg (Original) \\ 
FP: CLIPSeg (FP16) \\
RPE: CLIPSeg + Reusing Prompt Embeddings \\
FP-RPE: CLIPSeg (FP16) + Reusing Prompt Embeddings \\
HF: CLIPSeg (Hugging Face) \\
HF-RPE: CLIPSeg (Hugging Face) + Reusing Prompt Embeddings \\
FP-RPPE: CLIPSeg (FP16) + Reusing Prompt \& Positional Embeddings \\
FP-RPPET: CLIPSeg (FP16) + Reusing Prompt \& Positional Embeddings + TensorRT \\
FP-RPPETI: CLIPSeg (FP16) + Reusing Prompt \& Positional Embeddings + TensorRT + Input/Output Binding.}
\label{figure:freq}
\end{figure}

\subsection{Computational Efficiency}
\begin{figure}[htbp]
\centerline{\includegraphics[width =1\linewidth]{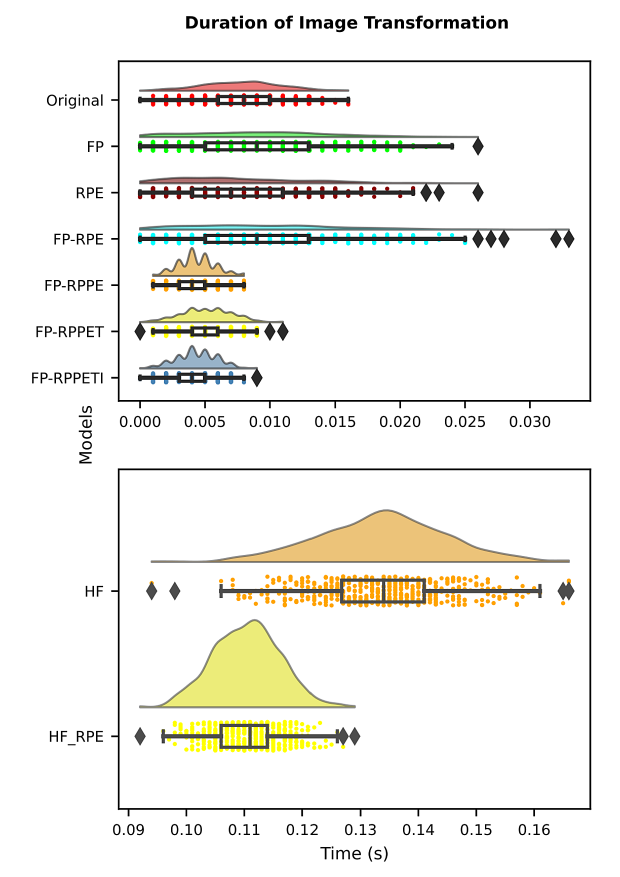}}
\caption{Duration of Image Transformation before Segmentation.}
\label{figure:transform}
\end{figure}

Through quantization of the model to FP16, we increased the model frequency from
1.81Hz to 4.98Hz (275.13\% increase, Fig.~\ref{figure:freq}). Additionally, by
reusing prompt and positional embeddings, the model frequency increased to
9.23Hz, which is 509.89\% increase compared to the original CLIPSeg. Additional
optimization through input/output binding and converting our ONNX
model to TensorRT led to an average of 927.41\% in improvement. We also analyzed the
effect of optimization on image transformation performed before segmentation and
noticed that reusing positional embeddings results in significant decrease in
image processing time (Fig.~\ref{figure:transform}), contributing to the faster
overall frequency.

\begin{figure}[htbp]
\centerline{\includegraphics[width =1\linewidth, trim = 0 3.9cm 0 3.9cm, clip]{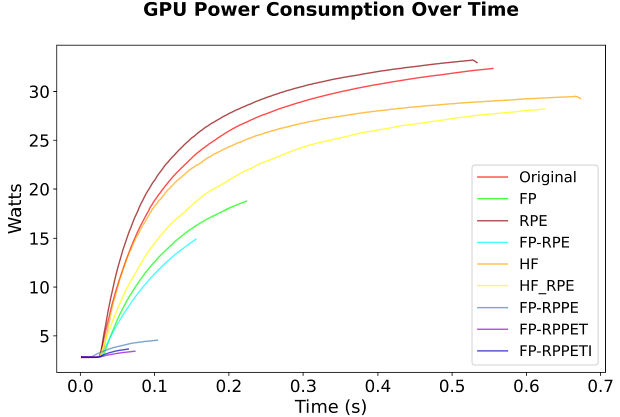}}
\caption{GPU Usage During Optimization.}
\label{figure:gpu_eff}
\end{figure}

\begin{figure}[htbp]
\centerline{\includegraphics[width =1\linewidth, trim = 0 3.7cm 0 3.65cm, clip]{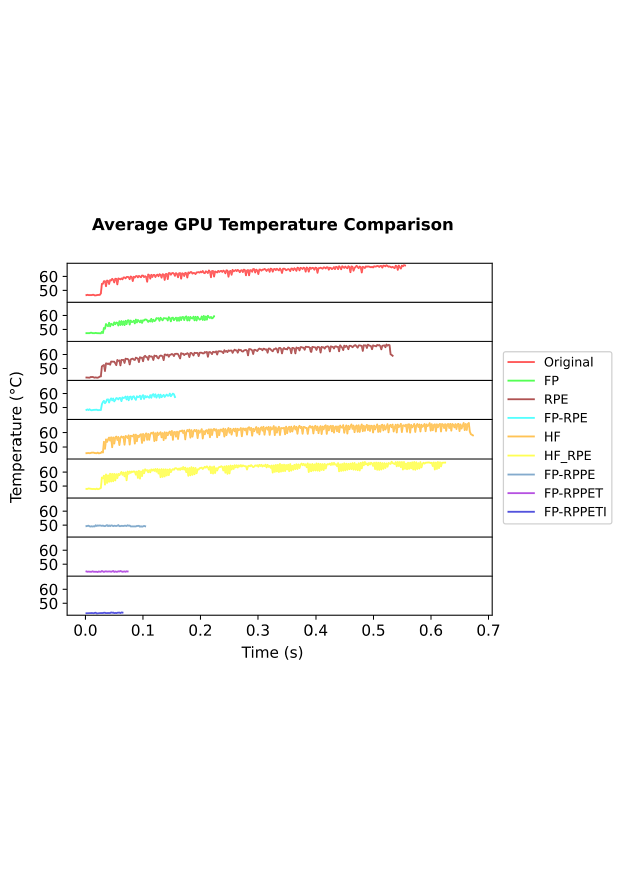}}
\caption{GPU Temperature Measurement During Optimization.}
\label{figure:gpu_ther}
\end{figure}

Using the NVIDIA Jetson's built-in Power GUI, we measured an approximate GPU
usage and temperature. Reusing of prompt and positional embedding showed
promising results in regards to GPU efficiency. Impressive reduction of GPU
utilization and temperature was observed by the reusing architecture as as shown
in figures~\ref{figure:gpu_eff},~\ref{figure:gpu_ther}. Overall, we observed
that GPU utilization and temperature dropped almost 10 times and 20\textcelsius,
respectively, comparing the original CLIPSeg (approximate maximum GPU
utilization at 33W and temperature at 68\textcelsius) with BlabberSeg
(approximate maximum GPU utilization at 3.6W and temperature at 45\textcelsius).
Note that the GPU measurements are an estimation, but they demonstrate the
efficiency of our optimization methods.

\subsection{Model Accuracy}
\begin{table*}[htbp] 
    \centering    
    \caption{Quantitative results of Comparison Among Original CLIPSeg and Optimized CLIPSeg Models for NVIDIA Jetson Orin AGX 64G.}
    \label{comparision}    
\renewcommand\arraystretch{1}
\begin{adjustbox}{width = 1\textwidth} 
\begin{tabular}{c|lccccc|}   
\textbf{Device} & 
\textbf{Models} &
\textbf{Mean Segmentation Duration (s)} &
\textbf{Mean Accuracy (\%)} &
\textbf{mIoU} \\
\hline
\multirow{9}*{\begin{tabular}[c]{@{}l@{}}NVIDIA \\ Jetson Orin \\ AGX 64G\end{tabular}} 
& Original & 0.553$\pm$0.004 & 100.0 (ground truth) & 100.0 (ground truth) \\
& FP & 0.201$\pm$0.019 & 99.5$\pm$0.29 & 99.6$\pm$0.36 \\
& HF & 0.677$\pm$0.032 & 100.0 (ground truth) & 100.0 (ground truth) \\
& HF-RPE & 0.617$\pm$0.012 & 100.0 (ground truth) & 100.0 (ground truth) \\
& RPE & 0.547$\pm$0.044 & 100.0 (ground truth) & 100.0 (ground truth)\\
& FP-RPE & 0.159$\pm$0.010 & 99.5$\pm$0.29 & 99.6$\pm$0.36 \\
& FP-RPPE & 0.108$\pm$0.009 & 99.5$\pm$0.01 & 97.7$\pm$1.49 \\
& FP-RPPET & 0.077$\pm$0.006 & 97.9$\pm$1.27 & 91.0$\pm$8.40 \\
& FP-RPPETI & \textbf{0.060}$\pm$0.009 & 97.9$\pm$1.27 & 91.0$\pm$8.40
\end{tabular}
\end{adjustbox}
\end{table*}

\begin{table*}[htbp] 
    \centering    
    \caption{Quantitative results of Comparison Among Original CLIPSeg and Optimized CLIPSeg Models for Other NVIDIA Jetson devices}
    \label{comparision2}    
\renewcommand\arraystretch{1}
\begin{adjustbox}{width = 1\textwidth} 
\begin{tabular}{c|lccccc|}   
\textbf{Device} & 
\textbf{Models} &
\textbf{Mean Segmentation Duration (s)} &
\textbf{Increase in Speed (\%)} &
\textbf{Hz} \\
\hline
\multirow{9}*{\begin{tabular}[c]{@{}l@{}}NVIDIA \\Jetson \\ Orin NX 16G \\ (Emulated)\end{tabular}} 
& Original & 1.490$\pm$0.067 & 100.0 (ground truth) & 0.67$\pm$0.0301 \\
& FP & 0.420$\pm$0.023 & 354.7$\pm$4.31 & 2.38$\pm$0.1296 \\
& HF & 1.537$\pm$0.040 & 97.0$\pm$0.62 & 0.65$\pm$0.0185 \\
& HF-RPE & 1.469$\pm$0.0474 & 101.5$\pm$4.27 & 0.68$\pm$0.1285 \\
& RPE & 1.364$\pm$0.035 & 109.2$\pm$0.55 & 0.73$\pm$0.0166 \\
& FP-RPE & 0.378$\pm$0.018 & 394.5$\pm$0.73 & 2.65$\pm$0.0220 \\
& FP-RPPE & 0.118$\pm$0.008 & 1258.2$\pm$18.95 & 8.44$\pm$0.5705 \\
& FP-RPPET & 0.094$\pm$0.009 & 1592.6$\pm$35.17 & 10.69$\pm$1.0588 \\
& FP-RPPETI & \textbf{0.068}$\pm$0.007 & \textbf{2205.0}$\pm$53.72 & \textbf{14.79}$\pm$1.6174 \\
\hline
\multirow{9}*{\begin{tabular}[c]{@{}l@{}}NVIDIA \\ Jetson Orin \\ Nano 8G \\ (Emulated)\end{tabular}} 
& Original & 2.135$\pm$0.033 & 100.0 (ground truth) & 0.47$\pm$0.0072 \\
& FP & 0.594$\pm$0.023 & 359.2$\pm$8.91 & 1.68$\pm$0.0643 \\
& HF & 2.184$\pm$0.017 & 97.7$\pm$2.35 & 0.46$\pm$0.0170 \\
& HF-RPE & 2.105$\pm$0.016 & 101.4$\pm$8.00 & 0.48$\pm$0.0577 \\
& RPE & 2.044$\pm$0.071 & 104.4$\pm$0.50 & 0.49$\pm$0.0036 \\
& FP-RPE & 0.553$\pm$0.018 & 386.2$\pm$0.50 & 1.81$\pm$0.0036 \\
& FP-RPPE & 0.150$\pm$0.009 & 1424.7$\pm$57.81 & 6.67$\pm$0.4171 \\
& FP-RPPET & 0.099$\pm$0.007 & 2147.0$\pm$97.29 & 10.06$\pm$0.7019 \\
& FP-RPPETI & \textbf{0.079}$\pm$0.008 & \textbf{2695.2}$\pm$166.29 & \textbf{12.63}$\pm$1.1997 \\
\hline
\multirow{9}*{\begin{tabular}[c]{@{}l@{}}NVIDIA \\ Jetson AGX \\Xavier 64G\end{tabular}} 
& Original & 1.570 $\pm$0.078 & 100.0 (ground truth) & 0.64$\pm$ 0.0317\\
& FP & 0.419$\pm$0.048 & 374.6$\pm$8.72 & 2.39$\pm$0.2767 \\
& HF & 1.615$\pm$0.051 & 97.2$\pm$0.16 & 0.62$\pm$0.0050 \\
& HF-RPE & 1.550$\pm$0.040 & 101.3$\pm$3.79 & 0.65$\pm$0.1204 \\
& RPE & 1.417$\pm$0.010 & 110.8$\pm$0.61 & 0.71$\pm$0.0194 \\
& FP-RPE & 0.356$\pm$0.015 & 441.1$\pm$0.52 & 2.81$\pm$0.0166 \\
& FP-RPPE & 0.112$\pm$0.021 & 1404.4$\pm$55.58 & 8.94$\pm$1.7636 \\
& FP-RPPET & 0.076$\pm$0.003 & 2055.5$\pm$13.63 & 13.09$\pm$0.4326 \\
& FP-RPPETI & \textbf{0.066}$\pm$0.008 & \textbf{2383.2}$\pm$59.78 & \textbf{15.18}$\pm$1.8970 \\
\hline
\multirow{5}*{\begin{tabular}[c]{@{}l@{}}Raspberry \\Pi 5\end{tabular}} 
& Original & 42.973$\pm$1.536 & 100.0 (ground truth) & 0.02$\pm$0.0008 \\
& HF & 27.641$\pm$0.888 & 155.5$\pm$0.49 & 0.04$\pm$0.0004 \\
& HF-RPE & 21.154$\pm$0.255 & 203.2$\pm$1.40 & 0.05$\pm$0.0012 \\
& RPE & 21.770$\pm$0.193 & 197.4$\pm$0.68 & 0.05$\pm$0.00006 \\
& RPPE & \textbf{6.096}$\pm$0.966 & \textbf{715.5}$\pm$33.03 & \textbf{0.17}$\pm$0.0275
\end{tabular}
\end{adjustbox}
\end{table*}

\begin{figure}[htbp]
\centerline{\includegraphics[width = 1\linewidth]{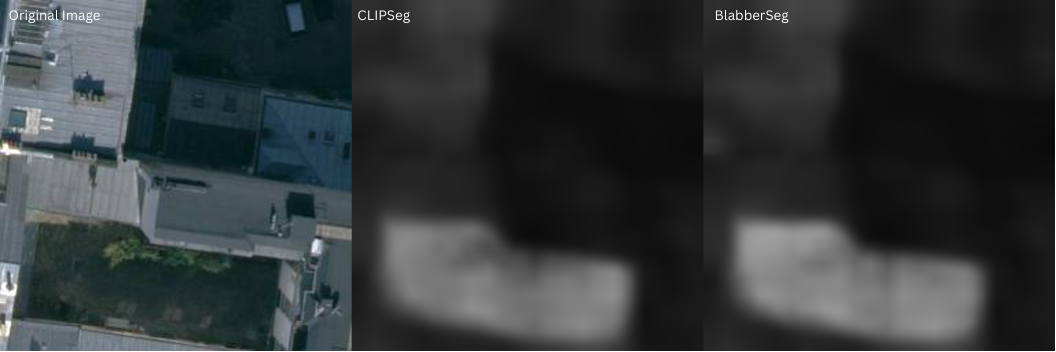}}
\caption{Difference in Segmentation: Original CLIPSeg (middle) vs BlabberSeg (right).}
\label{figure:diffSeg}
\end{figure}

We used the original CLIPSeg model accuracy and mIoU as ground truth and compared them with optimized models. Hugging Face models resulted in the same accuracy and mIoU since reusing prompt embeddings does not change anything on the models. As expected, quantization by converting the original CLIPSeg (FP32) to FP16 decreased mean accuracy and mIoU to 99.5 and 99.6, respectively (Table~\ref{comparision}). For the same reason as reusing prompt enbeddings, Input/Output binding did not decrease the accuracy and IoU as expected. Reusing positional embeddings decreased mean accuracy (99.5) and mIoU (97.5) (Table~\ref{comparision}) slightly, but major drop was from converting the model to TensorRT engine (mean accuracy: 97.9 and mIoU: 91.0) (Table~\ref{comparision}). As observed in~\ref{figure:diffSeg}, visual difference in segmentation is almost unrecognizable, despite significant gain in efficiency.
Overall, NVIDIA Jetson Orin AGX 64GB was able to boost its segmentation frequency from 1.81Hz to 16.78 Hz (Fig.~\ref{figure:freq}) with only marginal differences of 2.1\% in accuracy and 9\% mIoU (Table~\ref{comparision}). 

\subsection{Model Robustness}
\begin{figure*}[htbp]
\centerline{\includegraphics[width =1\linewidth, trim = 0 4.3cm 0 4.3cm, clip]{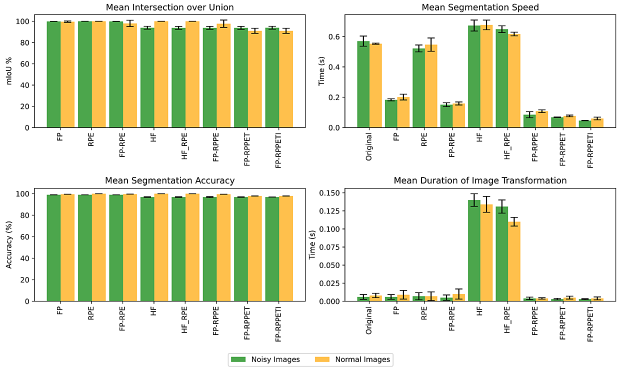}}
\caption{Comparison of Mean Intersection over Union (mIoU), mean duration of image transformation (seconds), mean segmentation accuracy (the ratio of the correctly segmented area over the ground truth (segmentation of the original CLIPSeg)) and speed between normal and noisy images. Plots for mIoU and mean segmentation accuracy does not include the orignal CLIPSeg (ground truth).}
\label{figure:noise}
\end{figure*}

Aerial images are often in low resolution due to natural condition, rapid motion and high altitude. To address the robustness of the model, additional experiments (100 images) conducted with the images used for our experiments~\cite{GoogleMaps2023}, but polluted with random noises. Compared to results generated using images that are not noisy, all metrics performed equally well with noisy images (Fig.~\ref{figure:noise}).

\subsection{Performance on Other CIoT Devices}
For other CIoT devices, we also observed promising increase in speed while keeping relatively high accuracy and mIoU as NVIDIA Jetson Orin AGX 64GB. NVIDIA Jetson Orin NX 16GB was able to increase its speed by 2205.01\% and perform segmentation at 14.79Hz, compared to the original CLIPSeg at 1.490Hz (Table~\ref{comparision2}). For NVIDIA Jetson Orin Nano 8G, we achieved 2695.19\% increase in speed which is significant and 12.63Hz, compared to 0.47Hz using the original CLIPSeg (Table~\ref{comparision2}). On Jetson AGX Xavier 64G, we observed 2383.23\% increase in segmentation speed and 15.18Hz. 

We also experimented with Raspberry Pi 5~\cite{rpi5} and measured its performance using only CPU. Due to the CPU's limitation, we measured all models with FP32 and without TensorRT and ONNX runtime. More specifically, CPU architectures are not designed to process FP16 operations or use TensorRT and ONNX enabled models. We observed better performance using the Hugging Face CLIPSeg model (0.04Hz) compared to the original CLIPSeg models (0.02Hz), which was not the case with Jetson devices (Table~\ref{comparision2}). However, their performance were similar after prompt reusing (0.05Hz). The possible reasoning behind the better performance on Hugging Face CLIPSeg model might be related to how it processes the prompts more efficiently using CPU compared to the original CLIPSeg, which might be more efficient in processing the prompts using GPU. Overall, BlabberSeg was able to boost its performance up to 0.17Hz per image segmentation, which is an increase of 715.54\% compared to the original CLIPSeg (0.02Hz). 

\section{Conclusion}
Although the objective of our work was to improve on computational latency, robustness on segmentation is paramount for successful UAV flight operations. By enabling CIoT devices with lightweight VLM aerial segmentation, our efforts reduce the gap of achieving close-to autonomous UAVs. With CIoT devices specifically tailored for AI-Enabled aerial computing, we expect to see more interest in the research of lightweight architectural design for foundation models in UAVs near future. Additionally, there is a need for rigorous tests on hardware deployments of CIoT devices equipped with foundation models to stress-test and validate these models to enable fault-tolerant usage in the setting of UAVs. 

\section{Acknowledgement}
We would like to highlight the support provided by National Research Council (NRC) Canada for their research guidance and funding, which made our work possible.

\bibliography{iot}
\bibliographystyle{IEEEtran}

\vskip -2\baselineskip plus -1fil 
\begin{IEEEbiography}[{\includegraphics[width=1in, height=1in]{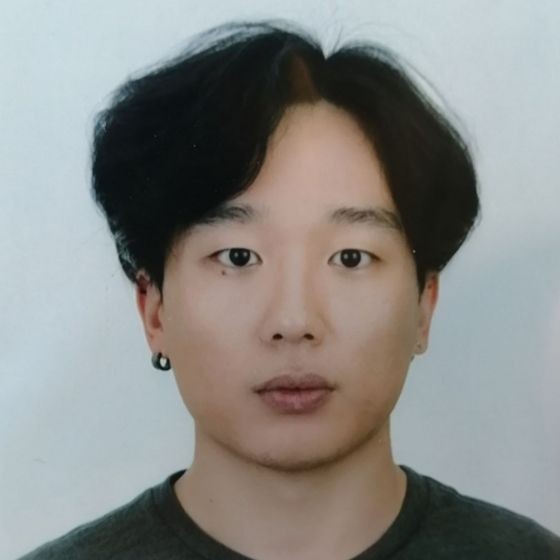}}]%
{Haechan Mark Bong}
(Student Member, IEEE) received his BSc. and MSc. from McGill University, Montréal, QC., Canada, in Computer Science (2018) and Animal Science (2022), respectively. He is currently pursuing his Ph.D. in robotics in Computer and Software Engineering Department in Polytechnique Montréal, QC., Canada, specializing in the application of Vision-Language models in robotics. His research interests are in robotics, LLMs, VLMs, MLLMs, CIoT devices and UAVs.
\end{IEEEbiography}
\vskip -3\baselineskip plus -1fil 
\begin{IEEEbiography}[{\includegraphics[width=1in, height=1in]{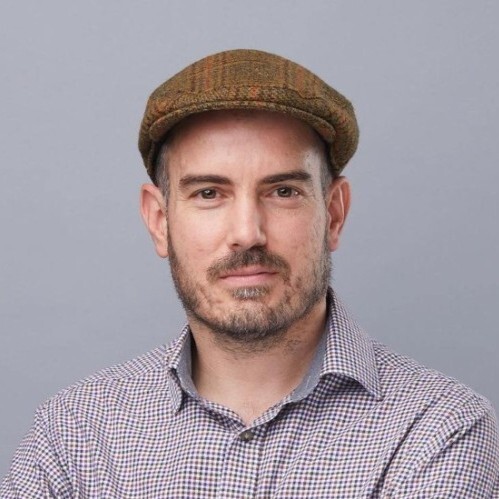}}]%
{Ricardo de Azambuja}
(Member, IEEE) received his B.Sc. (2006) and M.Sc. (2013) degrees in Electrical Engineering from the Federal University of Rio Grande do Sul (UFRGS), Porto Alegre, Brazil, and the Ph.D. degree in Computer Science (Spiking Neural Networks for Humanoid Robot Control) from the University of Plymouth, UK, in 2017. His research journey has spanned diverse areas, including wireless powered sensors, bio-inspired spiking neural networks, collaborative robotics, computer vision, and collision-resilient drones. Currently, he works as a Senior Automation Engineer at Questat.ca, where he is contributing to revolutionizing the biotech blood gases sensor sector.
\end{IEEEbiography}
\vskip -2\baselineskip plus -1fil 
\begin{IEEEbiography}[{\includegraphics[trim={0 10cm 0 0}, height=1in, clip, width=1in]{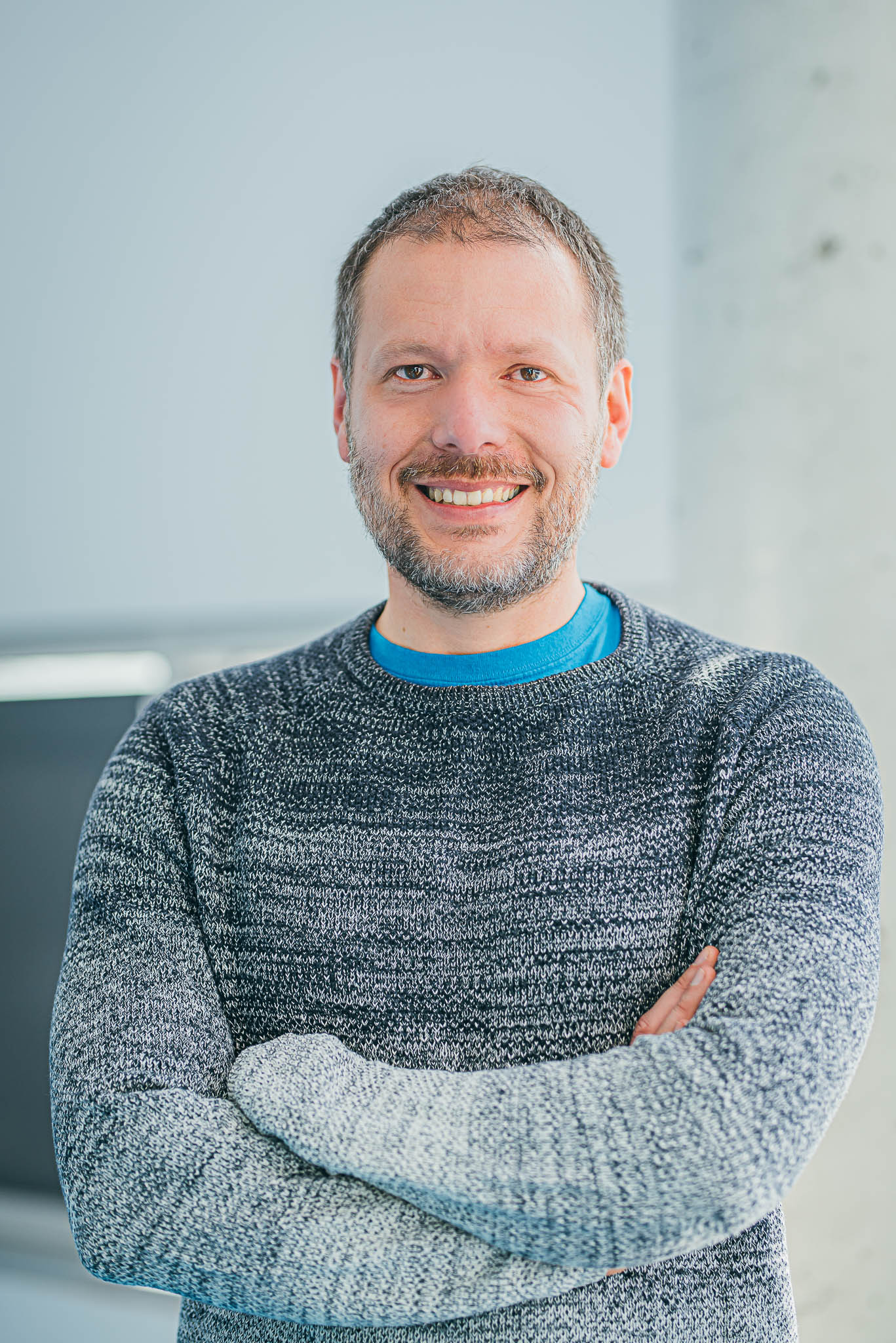}}]%
{Giovanni Beltrame}
(Senior Member, IEEE) received the Ph.D. degree in computer engineering from Politecnico di Milano, Milan, Italy, in 2006.
He worked as a Microelectronics Engineer with the European Space Agency, Paris, France, on a number of projects, spanning from radiation tolerant systems to computer-aided design. Since 2010, he has been the Professor with the Computer and Software Engineering Department, Polytechnique Montréal, Montréal, QC., Canada, where he directs the MIST Lab. He has authored or coauthored more than 100 papers in international journals and conferences. His research interests include modeling and design of embedded systems, artificial intelligence, and robotics. Dr. Beltrame was the recipient of more than 30 grants by government agencies and industry.
\end{IEEEbiography}
\end{document}